\RequirePackage{fix-cm}
\documentclass[twocolumn]{svjour3}          
\smartqed  
\usepackage{graphicx}
\usepackage{latexsym}
\usepackage{url}
\usepackage{xcolor}
\definecolor{newcolor}{rgb}{.8,.349,.1}
\usepackage{float}
	\definecolor{blue(munsell)}{rgb}{0.0, 0.5, 0.69}
\usepackage{caption}
\usepackage{multirow}
\usepackage[font={color=blue(munsell),bf},figurename=Fig.,labelfont={it}]{caption}
\usepackage{enumerate}
\usepackage{enumitem}
\usepackage{appendix}
 \usepackage{algorithm}
 \usepackage{algpseudocode}
 \usepackage{algcompatible}
\usepackage{hyperref}
\usepackage{amsmath}
\usepackage{colortbl}
\hypersetup{
     colorlinks = true,
     linkcolor = blue,
     anchorcolor = blue,
     citecolor = blue,
     filecolor = blue,
     urlcolor = blue
     }

\usepackage{array}
\usepackage{float}
\newcolumntype{A}{>{\centering\arraybackslash}m{1.2 cm}}
\newcolumntype{C}{>{\centering\arraybackslash}m{1.6 cm}}
\newcolumntype{E}{>{\centering\arraybackslash}m{1.4 cm}}
\begin{document}

\title{Deep learning based Hand gesture recognition system and design of a Human-Machine Interface 
}


\author{Abir  Sen         \and
        Tapas Kumar Mishra \and Ratnakar Dash 
}


\date{Received: date / Accepted: date}

\maketitle

\begin{abstract}
In this work, a real-time hand gesture recognition system-based human-computer interface (HCI) is presented. The system consists of six stages: (1) hand detection, (2) gesture segmentation, (3) use of five pre-trained convolutional neural network models (CNN) and vision transformer (ViT), (4) building an interactive human-machine interface (HMI), (5) development of a gesture-controlled virtual mouse, (6) use of Kalman filter to estimate the hand position, based on that the smoothness of the motion of pointer is improved. In our work, five pre-trained CNN (VGG16, VGG19, ResNet50, ResNet101, and Inception-V1) models and ViT have been employed to classify hand gesture images. Two multi-class datasets (one public and one custom) have been used to validate the models. Considering the model's performances, it is observed that Inception-V1 has significantly shown a better classification performance compared to the other four CNN models and ViT in terms of accuracy, precision, recall, and F-score values. We have also expanded this system to control some desktop applications (such as VLC player, audio player, file management, playing 2D Super-Mario-Bros game, etc.) with different customized gesture commands in real-time scenarios. The average speed of this system has reached 25 fps (frames per second), which meets the requirements for the real-time scenario. Performance of the proposed gesture control system obtained the average response time in milisecond for each control which makes it suitable for real-time. This model (prototype) will benefit physically disabled people interacting with desktops.


\end{abstract}

\section{Introduction}
\label{intro}

HCI has become a part of our daily life, such as for entertainment purposes or for getting in touch with the assistive interface. With the growth of computer vision, hand gesture recognition has become a trendy field for interaction between man and machine without physically touching any devices. Imagine a scenario such as smart home automation, where a user uses hand gestures to control devices like TV, music player control, wifi turn off/on, light on/off, and various applications. Another scenario is desktop application controls using hand gestures like VLC player control, audio player control, or 2D-game controls and the development of gesture-controlled virtual mouse and keyboard. So in both cases, users can control applications without physically contacting them, which is beneficial for physically impaired and older people. There are two types of hand gesture recognition: (1) wearable gloves based \cite{berezhnoy2018hand,abhishek2016glove}, and (2) vision-based \cite{7379635,al2022structured}. The disadvantage of the first method is that it is expensive, requires wearing on hand to recognize gestures, and is also unstable in some environments. The second method is based on image processing, where the pipeline of this methodology is followed as: capturing an image using the webcam, segmentation, feature extractions, and classifications of gestures. But challenges exist in the development of a robust hand-gesture recognition system. Though the performance has been enhanced by using advanced depth-sensing cameras like Microsoft Kinect or Intel Real-Sense, they are relatively costly, so it is still an issue. The second challenge dealt by this project is building a human-machine interface with gesture commands. The third issue resolved by our work is the implementation of the gesture-controlled virtual mouse in real-time and the movement of the mouse cursor stably and smoothly. We know that our hand is not always stable for controlling the mouse cursor's movement, especially while changing the gestures to understand different commands. So this work tries to address the above-mentioned issues by proposing a vision-based hand gesture recognition system using CNN models and ViT, followed by extending our work to develop a real-time gesture-controlled human-machine interface. To make our system more convenient and user-friendly, we have also developed a virtual mouse with the help of gesture commands. The significant contributions of this article are summarized as follows:

\begin{enumerate}[label=\Roman*.] 
\item A low-cost vision-based hand gesture recognition system has been illustrated using five pre-trained CNN models and ViT. We have compared the performance among all considered CNN models and ViT. Then the best model has been selected for the real-time inference task.
    \item  An interactive human-machine interface using customized hand-gesture commands has been built to control different applications in real-time.
    
\item Different events of a physical mouse have been triggered with customized hand gesture commands to build a virtual mouse to make the HCI interface more convenient.

\item The Kalman filter has been employed to estimate and update the mouse cursor's position to enhance the smoothness of the cursor's motion and prevent jumping of the cursor on the whole screen from one point to another. 
\end{enumerate}

\hspace{-0.6cm} The remainder paper is structured as follows. In Section \ref{literature}, literature works related to different hand gesture recognition are illustrated. The proposed methodology is discussed in Section \ref{proposed}. Section \ref{experiment} deals with experimental details, dataset description, comparative analysis, and the details of the gesture-controlled human-machine interface followed by real-time performance analysis. Section \ref{discussion} describes the discussion portion. Finally, Section \ref{conclusion} includes the summary of the conclusion and future works.

\vspace{-0.4 cm}
\section{\label{literature} Literature survey}
In the past few years, with the rapid growth of computer vision, hand gesture recognition has facilitated the field of human-machine interaction. Computer vision researchers mainly use the web camera and hand gestures as inputs to the gesture recognition systems. But while building a low-cost human-machine interface in real-time, some challenges exist, such as background removal, gesture segmentation, and classifying gesture images in messy backgrounds. So the classification of gestures is a crucial task in developing a gesture-controlled HCI. Various state-of-art works for hand gesture recognition were introduced in the last decade. Some used classification techniques are artificial neural network (ANN) algorithms, Hidden Markov model (HMM), or support vector machine (SVM). For instance, Mantecon et al. \cite{mantecon2016hand} proposed a hand gesture recognition system using infrared images captured by Leap motion controller device that produces image descriptors without using any segmentation phase. Next, the descriptors were dimensionally reduced and fed into SVM for final classification. Huang et al. \cite{huang2011gabor} developed a novel hand gesture recognition system by using the Gabor filter for feature extraction tasks and SVM for the classification of hand gesture images. Singha et al. \cite{singha2018dynamic} developed a dynamic gesture recognition system in the presence of cluttered background. Their proposed scheme consists of three stages: (1) hand portion detection by three-frame difference and skin-filter technique; (2) feature extraction; (3) feeding the features into ANN, SVM, and KNN classifiers, followed by combining these models to produce a new fusion model for final classification. 
 In \cite{yang2012dynamic}, skin color segmentation was done to separate the hand region from image sequences after collecting the gesture images from the web camera. HMM was employed on hand feature vectors to classify complex hand gestures. So in various cases, the pipeline of a vision-based hand gesture recognition system is limited to detection, segmentation, and classification. \\ But recently, with the rapid growth of computer vision and deep learning, the CNN model has become a popular choice for feature extraction and classification task. Recent research shows that gesture classification with deep learning algorithms has produced better results than earlier proposed schemes. For instance, Yingxin et al. \cite{yingxin2016robust} developed a hand gesture recognition system, where the canny edge detection technique was employed for pre-processing and CNN for both feature extraction and classification task. Oyebade et al. \cite{oyedotun2017deep} presented a vision-based hand gesture recognition for recognizing American Sign language. Their proposed scheme has two stages: (1) segmentation of gesture portion after applying binary thresholding and (2) using CNN architecture to predict the final class. In \cite{fang2019gesture}, the authors developed a hand gesture recognition system based on CNN. Later deep convolution generative adversarial networks (DCGAN) was used to avoid the overfitting problem. In \cite{adithya2020deep}, the experiment was carried out on publicly available datasets (NUS and American finger-spelling dataset). Here CNN was employed for classifying the static gesture images. Recently, Sen et al. \cite{sen2022novel} developed a hand gesture recognition system, which had three stages; (1) gesture detection by binary thresholding, (2) segmentation of gesture portion, (3) training of three custom CNN classifiers in parallel, followed by calculating the output scores of these models to build the ensemble model for final prediction.  \\   
Recently, Vision Transformer \cite{dosovitskiy2020image} has exhibited excellent results for image classification tasks. The performance of the vision transformer is completely based on the transformer model, which is widely used in natural language processing (NLP). However, there are fewer research papers on vision transformer for the vision-based gesture recognition task. For example, In \cite{godoy2022electromyography}, the authors published an article titled `Electromyograph based, robust hand motion classification employing temporal multi-channel vision transformers', where they have proposed a multichannel based vision transformer to perform the surface-electromyograph (sEMG) based hand motion classification task. In \cite{montazerin2022vit}, the authors have applied ViT to classify sEMG signals and have shown some fruitful results. \\
Furthermore, to build a human-machine interface, a gesture-controlled virtual mouse is necessary that eliminates the physical device dependency. Some literature studies have shown fruitful results regarding the performance of the gesture-controlled virtual mouse. For example, In \cite{shibly2019design}, Shibly et al. have developed a gesture-controlled virtual mouse by using web-camera captured frames processing and color detection method for gesture portion detection. In the article \cite{tsai2020design}, the authors proposed a low-cost gesture recognition system in a real-time scenario, which has three stages; (1) skin segmentation, (2) moving gesture detection by motion-based background separation, (3) separate of arm portion for isolating the palm area followed by then analyze the hand portion using the convex hull method. They also developed a virtual mouse by triggering all functions of a physical mouse with different customized gestures, and their proposed scheme achieved a recognition rate of more than 83\% during inference time.\\
But, in a real-time scenario, controlling a virtual mouse is a very challenging task, as a slight hand movement can lead to the jump of the virtual mouse cursor over the screen. So it is necessary to be concerned about the smoothness of the motion of the gesture-controlled mouse cursor. Literature survey has shown very few promising results about the smooth movement of the gesture-controlled mouse-cursor \cite{xu2017real}. In this work, we attempt to develop a low-cost hand gesture recognition system along with a virtual mouse for real-time.
\vspace{-0.4 cm}
 \section{\label{proposed} Proposed methodology}
\begin{figure*}[!ht]
\centering
\includegraphics[width=\textwidth,height=5 cm]{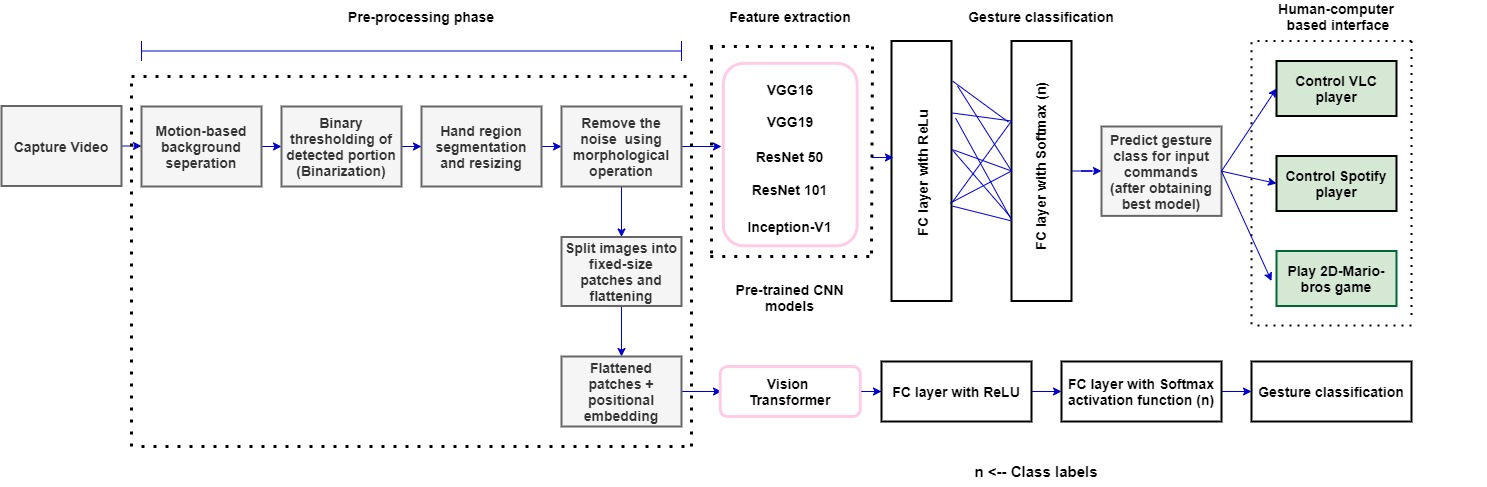}
\caption{\label{model_diagram} Illustration of our proposed scheme.}
\end{figure*}

This section gives a detailed description of the proposed methodology. Initially, some preprocessing, such as gesture detection by motion-based background separation, binary thresholding, contour portion selection, hand segmentation, resizing, and use of morphological filter for noise removal, are carried out. Then the resized images are passed into five pre-trained CNN models (VGG16, VGG19, ResNet50, ResNet101, and Inception-V1) for training them separately to obtain the final predicted gesture class. We have also used the vision transformer (ViT) model to classify gesture images. In the final part, we have extended our work by building a human-machine interface that utilizes gesture labels as input commands to control some desktop applications in real-time scenario. The overview of our proposed framework is depicted in Fig. \ref{model_diagram}.
\subsection{\label{preprocessing} Pre-processing phase}
\begin{figure*}[!ht]
\centering
\includegraphics[width=\textwidth,height=4.5cm]{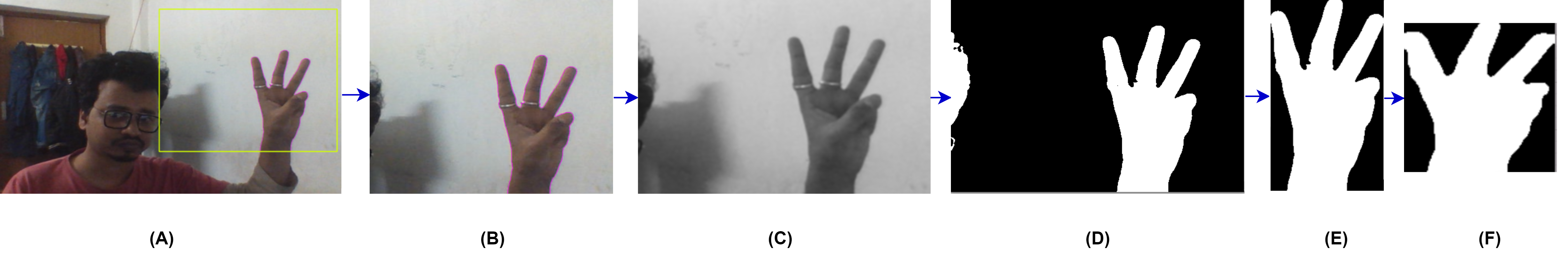}

\caption{\label{preprocessing_phase} The whole pre-processing phase of our proposed work (A) Input frame, (B) Gesture region detection, (C) conversion into gray scale, (D) Binary thresholding, (E) Extraction of hand region, (F) Applying of morphological operation for noise removal of segmented images after resizing.}
\end{figure*}

Image preprocessing is one of the most significant and vital steps to improve model performance. This phase comprises some stages, such as gesture detection by motion-based background separation, binary thresholding of detected hand region, segmentation by using contour region selection, then resizing of segmented images followed by applying the morphological transformation to remove noise. Next, the resized images are fed into five considered CNN models and ViT (as fixed-length image patches) for the classification task. All the steps are discussed below, and the entire pre-processing phase is shown in Fig. \ref{preprocessing_phase}.

\subsubsection{Gesture detection by motion-based background separation}
Motion-based background separation aims to identify the moving object from the continuous frame sequences. Initially, we consider the background, which contains no movement of hands. With the movement of hand, the hand portion is obtained by subtracting the current frame from the background image frame. During this stage, skin color is used to  separate the hand portion from the other moving objects in the frame. The skin color is obtained with the help of HSV (hue, saturation, and value) color model \cite{chen2014real}. The stage of detecting the hand region is illustrated in Fig. \ref{preprocessing_phase} (B). 

\subsubsection{Binary thresholding}
\begin{figure}[!ht]
\centering
\includegraphics[width=8cm,height=3.5cm]{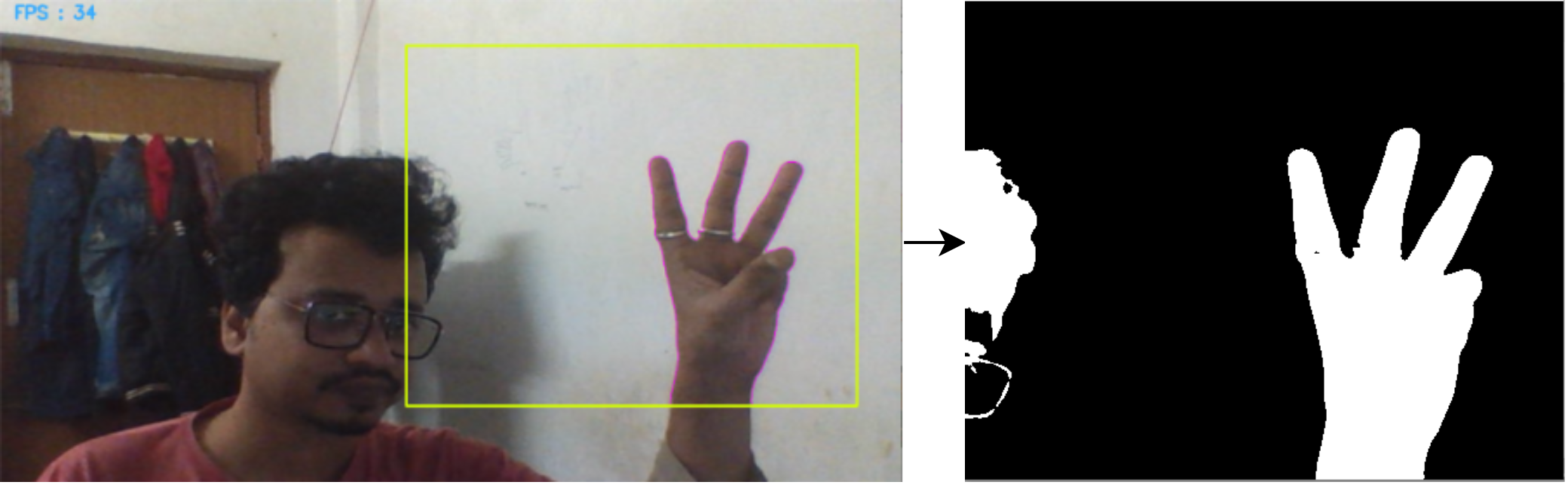}
\caption{\label{hand_detection}Hand portion detection in real-time.}
\end{figure}

 After gesture detection, we have applied a binary thresholding technique to segment between the background and foreground portions, making the background portion black and the detected hand portion white in color. An example of detected hand portion in real-time scenario has been displayed in Fig. \ref{hand_detection}.

\subsubsection{Contour region selection and segmentation of hand portion}
\begin{figure}[!ht]
\centering
\includegraphics[width=8cm,height=3.5cm]{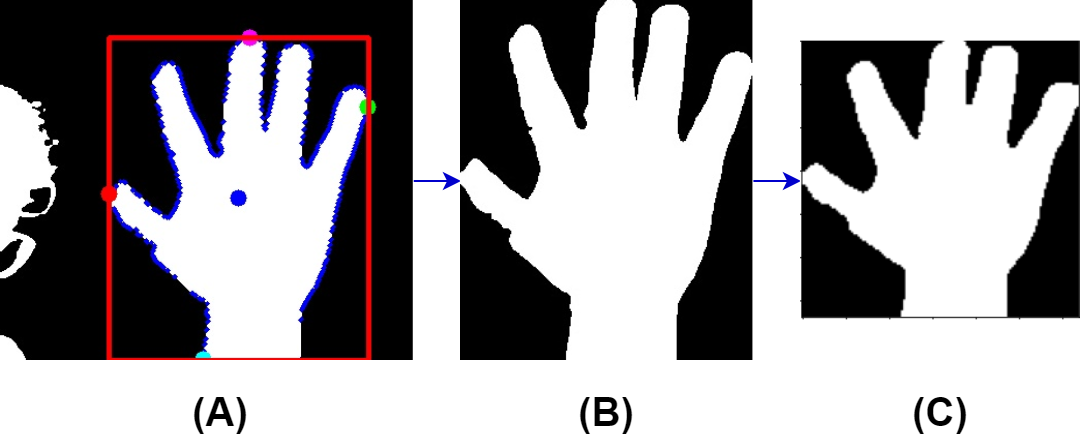}
\caption{\label{segmented}Hand region segmentation procedure (A) Contour portion bounded by rectangle box, here blue point represents the hand center (B) segmented hand region, (C) resized segmented hand portion.}
\end{figure}

        Contour region mainly indicates the outline/boundary of an object present in an image. To segment the hand region, firstly, we need to get the contour region of the detected portion. Due to cluttered background or low-light situations, there might be multiple holes/contour portions in the detected portion. But the contour portion, with the largest area, is considered the hand region. The hand region is extracted from the arm by using the distance transformation \cite{chen2014real}. Using this method, the distances between the pixels and the nearest boundary pixels are calculated \cite{xu2017real,chen2014real}. In a binary image's distance transformation matrix, the pixel with the largest distance is considered as palm or hand center. Tables \ref{binary}, \ref{distance} show the binary image and its distance transformation representation in matrix format. In Table \ref{distance}, it is observed that the largest distance is 5 (marked with light blue color), representing the hand /palm center. Hence the palm radius is estimated by calculating the minimum distance between the hand center and the point outside the contour region. Thus the hand region is separated from the arm based on wrist location. The whole diagram of contour portion selection to the segmentation phase has been shown in Fig. \ref{segmented}.

\begin{table}[!ht]
\caption{\label{binary}Representation of binary image in matrix format}
\centering
\begin{tabular}{|l|l|l|l|l|l|l|l|l|l|l|} 
\hline
0 & 0 & 0 & 0 & 0 & 0 & 0 & 0 & 0 & 0 & 0  \\ 
\hline
0 & 1 & 1 & 1 & 1 & 1 & 1 & 1 & 1 & 1 & 0  \\ 
\hline
0 & 1 & 1 & 1 & 1 & 1 & 1 & 1 & 1 & 1 & 0  \\ 
\hline
0 & 1 & 1 & 1 & 1 & 1 & 1 & 1 & 1 & 1 & 0  \\ 
\hline
0 & 1 & 1 & 1 & 1 & 1 & 1 & 1 & 1 & 1 & 0  \\ 
\hline
0 & 1 & 1 & 1 & 1 & 1 & 1 & 1 & 1 & 1 & 0  \\ 
\hline
0 & 1 & 1 & 1 & 1 & 1 & 1 & 1 & 1 & 1 & 0  \\ 
\hline
0 & 1 & 1 & 1 & 1 & 1 & 1 & 1 & 1 & 1 & 0  \\ 
\hline
0 & 1 & 1 & 1 & 1 & 1 & 1 & 1 & 1 & 1 & 0  \\ 
\hline
0 & 1 & 1 & 1 & 1 & 1 & 1 & 1 & 1 & 1 & 0  \\ 
\hline
0 & 0 & 0 & 0 & 0 & 0 & 0 & 0 & 0 & 0 & 0  \\
\hline
\end{tabular}
\end{table}

\begin{table}[!ht]
\caption{\label{distance}Distance transform matrix representation of the binary image}
\centering
\begin{tabular}{|l|l|l|l|l|l|l|l|l|l|l|} 
\hline
0 & 0 & 0 & 0 & 0 & 0                                                             & 0 & 0 & 0 & 0 & 0  \\ 
\hline
0 & 1 & 1 & 1 & 1 & 1                                                             & 1 & 1 & 1 & 1 & 0  \\ 
\hline
0 & 1 & 2 & 2 & 2 & 2                                                             & 2 & 2 & 2 & 1 & 0  \\ 
\hline
0 & 1 & 2 & 3 & 3 & 3                                                             & 3 & 3 & 2 & 1 & 0  \\ 
\hline
0 & 1 & 2 & 3 & 4 & 4                                                             & 4 & 3 & 2 & 1 & 0  \\ 
\hline
0 & 1 & 2 & 3 & 4 & {\cellcolor[rgb]{0,0.588,0.941}}\textcolor[rgb]{0.502,0,0}{5} & 4 & 3 & 2 & 1 & 0  \\ 
\hline
0 & 1 & 2 & 3 & 4 & 4                                                             & 4 & 3 & 2 & 1 & 0  \\ 
\hline
0 & 1 & 2 & 3 & 3 & 3                                                             & 3 & 3 & 2 & 1 & 0  \\ 
\hline
0 & 1 & 2 & 2 & 2 & 2                                                             & 2 & 2 & 2 & 1 & 0  \\ 
\hline
0 & 1 & 1 & 1 & 1 & 1                                                             & 1 & 1 & 1 & 1 & 0  \\ 
\hline
0 & 0 & 0 & 0 & 0 & 0                                                             & 0 & 0 & 0 & 0 & 0  \\
\hline
\end{tabular}
\end{table}

\subsubsection{Resizing of segmented images}
\begin{figure}[!ht]
\centering
\includegraphics[width=6.5cm,height=4.5cm]{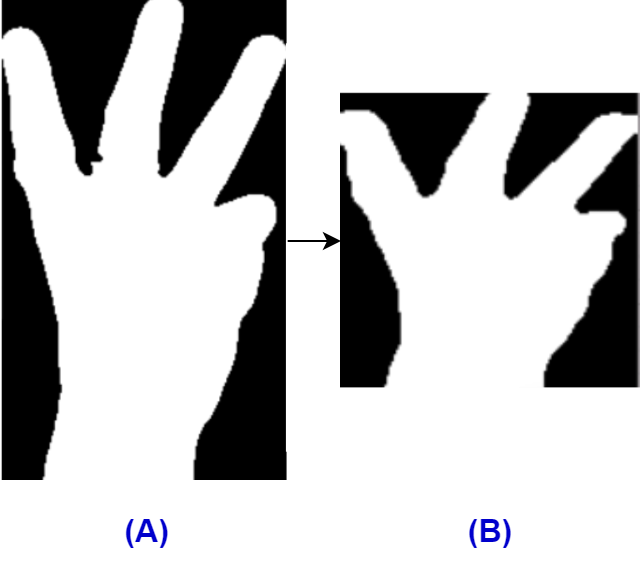}
\caption{\label{resizing}(A) Segmented gesture region (B) resizing of segmented hand portion into aspect ratio of $(64 \times 64 )$.}
\end{figure}

Image resizing is essential in the pre-preprocessing phase. In our experiment, the segmented images have been resized into the resolution of $(64 \times 64) $ and $(128 \times 128)$, as training with larger images consumes more time and computation cost than smaller images. Fig. \ref{resizing}, shows the segmented region and resizing of the hand portion.

\vspace{-0.1cm}
\subsubsection{Noise removal using morphological operation technique}
Morphological technique \cite{jamil2008noise} is a noise removal technique used to exclude the noise and enhance the appearance of the binary images. This technique comprises two phases, (1) dilation and (2) erosion. In our experiment, this technique is used to remove the holes and unwanted parts from the segmented gesture portion after resizing phase.
\vspace{-0.2cm}

\subsection{\label{transfer_learning} Feature extraction and classification using pre-trained CNN models and ViT} Building CNN models with millions of parameters from scratch is very time-consuming, with higher computation costs. So to overcome these issues, the transfer learning method has been employed. It is a very effective method to use CNN models for solving image-classification-related problems, where pre-trained models are reused to solve a new task. It is faster and requires less number of resources compared to CNN models built from scratch. In our work, some popular CNN models such as; VGG16 \cite{simonyan2014very}, VGG19 \cite{simonyan2014very}, ResNet50 \cite{he2016deep}, ResNet101, Inception-V1 \cite{szegedy2015going} have been explored for feature extraction task on multi-class hand gesture datasets using the transfer-learning approach. To obtain the predicted output, we have replaced each network's final fully connected (FC) dense layer with a new fully connected layer containing $n$ number of nodes, where $n$ denotes the number of classes present in
two considered datasets.\\
Recently, Vision transformer \cite{dosovitskiy2020image} has become a trendy topic, inspired by the concept of transformers and attention mechanism. It has been extensively used in image classification tasks. In \cite{dosovitskiy2020image}, the authors directly applied a Transformer to images by splitting the image into fixed-sized patches, then fed into the Transformer the sequence of embeddings for those patches. The image patches were treated as tokens in NLP applications. They also showed that vision transformer outperformed CNNs while trained on ImageNet dataset. So in our experiment, we will exploit vision transformer for gesture classification task. To obtain the predicted class, the final fully connected (FC) dense layer of this architecture is replaced with $n$ number of nodes, where $n$ represents the number of classes in two used datasets.
The details of these pre-trained CNN architectures and vision transformer are illustrated in section Appendix \ref{appendix-1}.
\begin{figure*}[!ht]
\centering
\includegraphics[width=\textwidth,height=9cm]{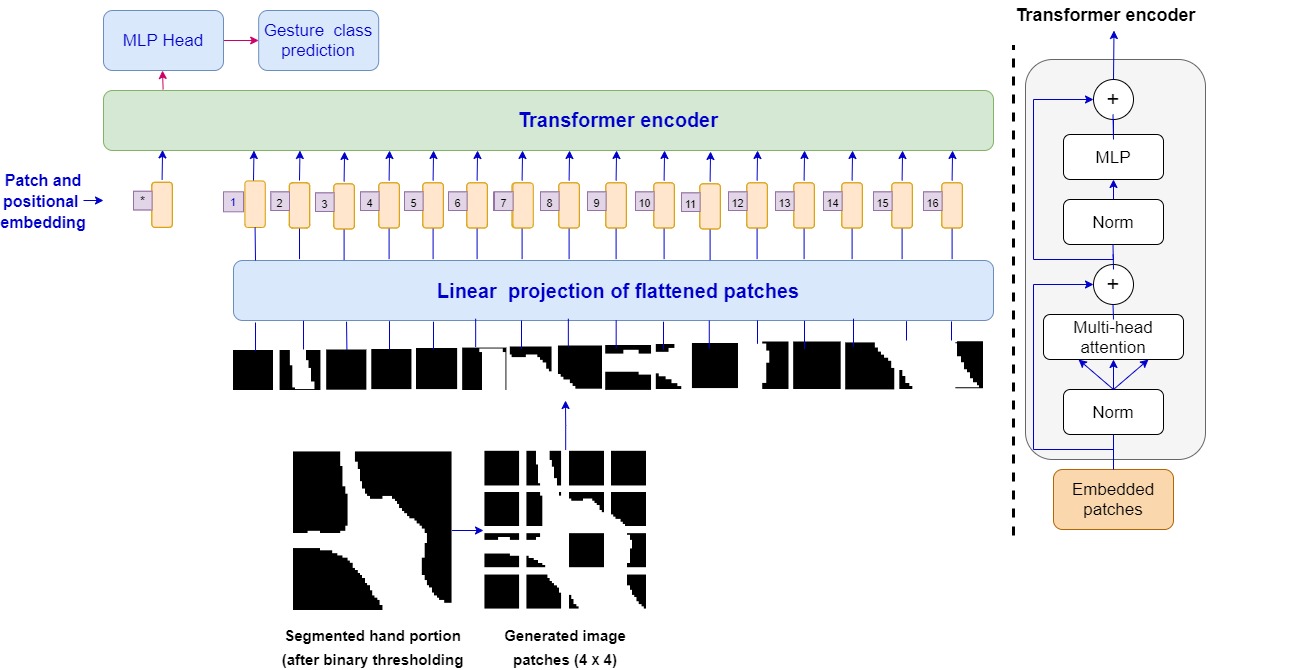}
\caption{\label{vision} Vision Transformer (ViT) architecture.}
\end{figure*}

\subsection{\label{Kalman_filter}Introducing of Kalman filter}
Kalman filtering \cite{Bang18} is one type of recursive predictive
algorithm that provides the estimation of the state at time $t$ in
a linear dynamic system in a continuous manner. 
The equations for state space model are:
\begin{equation}
x_{t+1}=A_{t}x_{t} + B_{t}u_{t} + \epsilon_{t} 
\end{equation}
\vspace{-0.6cm}
\begin{equation}
z_{t}=H_{t}x_{t}+ \beta_{t}
\end{equation}
where $x_{t}$ $A_{t}$, $B_{t}$, $u_{t}$, $z_{t}$, $H_{t}$ are the current state vector, state transition matrix, control input matrix, the measured state vector, and the observation matrix / transformation matrix respectively. Whereas $\epsilon_{t}$ and $\beta_{t}$ represent the process noise vector.\\
This filter comprises of two stages, (1) state prediction, (2) measurement update. In the first phase, it estimates the state in the system and in the second phase, estimated state is corrected resulting the error minimization in the covariance matrix.

 In our work, Kalman filter has been employed to enhance the smoothness of mouse-cursor movement. The use of Kalman filter is demonstrated in section \ref{mouse-cursor-kalman}.

\section{\label{experiment} Experiments and results}

To validate the proposed scheme, several experiments are carried out using  one publicly  available hand gesture  image datasets \cite{mantecon2016hand} (termed as Dataset-A), one self-constructed  dataset (termed as Dataset-B).
\subsection{\label{dataset_desciption}Datasets description}
\begin{figure}[!ht]
\centering
\includegraphics[width=8cm,height=4cm]{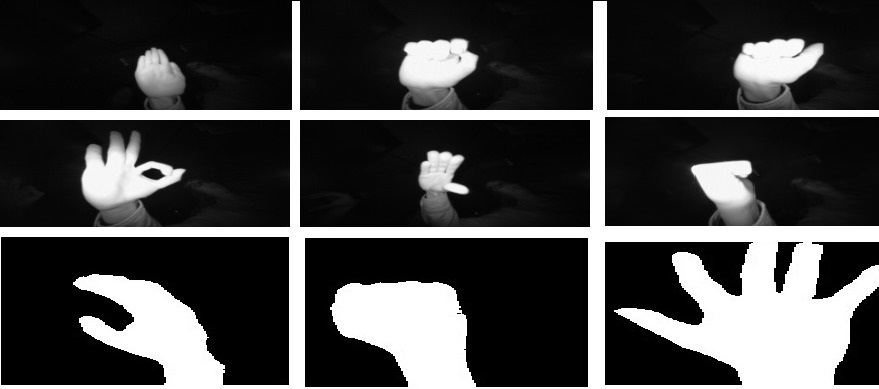}
\caption{\label{d1-sample} Samples images from Dataset-A and Dataset-B.}
\end{figure}

 Dataset-A comprises 20,000 gesture images of ten different gesture classes, such as: Fist, Fist move, Thumb, Index, Ok etc. The images of this dataset have been collected from ten  different  participants (five women and five  men), where each class label has 200 images. The images are having dimension of $(640 \times 240)$ pixels. 
 We have created our customized dataset (termed Dataset-B) containing 20,000 binary images collected by using the 16-megapixel USB webcam camera. This dataset contains 14 different gesture class labels collected by four participants (one woman and three men). Some samples of Dataset-A and Dataset-B have been illustrated in Fig. \ref{d1-sample}. The details of the two datasets have been tabulated in Table \ref{dataset}.
\hspace{-0.6cm}
\begin{table}[!ht]
\caption{\label{dataset}Dataset description}

\centering
\begin{tabular}{|C|C|C|C|}
\hline
Datasets  & Number of people performed & No of gesture label & Total no of images \\ \hline
Dataset-A & 10                         & 10                  & 20,000             \\ 
\hline
Dataset-B & 4                          & 14                  & 20,000             \\ \hline
\end{tabular}
\end{table}
\subsection{Dataset splitting}
To perform all experiments, the datasets are randomly split into 80:20 ratio for training and testing. Furthermore, for validation purposes, training sets have been split into 75\% for training and 25\% for validation. Now the percentage of training, validation and testing sets in the dataset will be 60\%, 20\%, and 20\%, respectively. The training samples are used to fit the model, validation set is used to evaluate the performance of each model for hyper-parameter tuning or to select the best model out of different models. The remaining testing sets are used to evaluate the performance of our proposed model.

\subsection{\label{experiment_details}Experimental setup \& details}
All experiments were carried out in Google Colab pro-environment, providing online cloud services with a Tesla P100 Graphical Processing Unit (GPU), Central Processing Unit (CPU), and 26 GB RAM. The all experiments carried out are divided 
into two parts: (1) `Experiment-1', (2) `Experiment-2'. Dataset-A has been considered for `Experiment-1' to solve the 10-class classification task. `Experiment-2' deals with solving 14-class classification task, which was carried out on Dataset-B.

In our experiments, after preprocessing phase (mentioned in section \ref{preprocessing}), the segmented gesture images are resized into dimensions of $(64\times 64)$ and $(128 \times 128)$ pixels followed by morphological transformation \cite{jamil2008noise} to exclude the noisy and redundant parts. Then they are fed into
five pre-trained models, such as VGG16, VGG19, ResNet50, ResNet101, and InceptionV1, for the feature extraction task, followed by adding a dense layer (as output layer) containing nodes according to the number of classes present in Dataset-A, Dataset-B, respectively, to predict the final gesture class label. While training the models, the softmax activation function is used to find the probability values of the last output layers, where the maximum probability value decides the final gesture class label. Fine-tuning is done using the Adam optimizer, where the $\beta_{1}$ and $\beta_{2}$ are 0.9 and 0.99, respectively. We have set the initial learning rate as 0.0001 later; it has been multiplied by 10 after ten epochs. The batch size and the number of epochs/iterations for all experiments are fixed as 64 and 30, respectively. \\
In the case of Vision Transformer, the segmented (after binary thresholding) images are split 
into the sequence of patches, and the size of each patch is fixed to ($6 \times 6 $) and flattened. Therefore each flattened patch is linearly projected, called ``patch embedding". The position embeddings are linearly added to keep the information of the sequence of image patches. Next, the sequences of vector images are fed into the transformer encoder as inputs. Finally, the last layer of MLP (added on top of the transformer encoder) predicts the gesture class. The entire structure of the vision transformer has been depicted in Fig. \ref{vision}.
The display of the  hyper-parameter setting for both experiments (using CNN models and ViT) is tabulated in Table \ref{parameter_value}.

\begin{table}[!ht]
 \caption{\label{parameter_value} Display of Hyper-parameter settings.}
\begin{center}
\resizebox*{0.505\textwidth}{!} {

\begin{tabular}{|c|c|}

\hline
\multicolumn{2}{|c|}{{\cellcolor[rgb]{0.851,0.8,0.878}}\textbf{Pre-trained CNN models}}   \\ 
\hline
Used parameters      & Values                     \\ \hline

Batch Size     & 64                       \\ \hline

Number of iterations         & 30                        \\ \hline

Learning rate  &  0.0001, 0.001 after 10 epochs                    \\ \hline
Pooling size    & ($2\times 2$)                  \\ 
\hline

Optimizer      & Adam                      \\ \hline

Error function      & Categorical cross entropy                      \\ \hline

Activation function      & ReLU, Softmax                    

 \\ 
\hline
\multicolumn{2}{c}{{\cellcolor[rgb]{0.851,0.8,0.878}} \textbf{Vision Transformer}}  

  \\ \hline

Batch Size                      & 64                                                                                                                             \\ \hline
Learning rate~                  & 0.0001                                                                                             \\ 
\hline
Patch size                      & ($6 \times 6$)~                                                                                                                       \\ 
\hline
Projection dimension~           & 64                                                                                                                             \\ 
\hline
Number of attention heads~      & 4                                                                                                                              \\ 
\hline
Number of transformer layers~ ~ & 8                                                                                                                              \\ 
\hline
Activation function             & GeLu                                                                                                                           \\ 
\hline
Droput rate                     & 0.5                                                                                                                            \\ 
\hline
MLP head size unit              & {[}2048,1024]                                                                                                                  \\
\hline

\end{tabular}
  }
\end{center}
\end{table}

\subsection{Results}
In this section, we have discussed various results obtained from our two experiments performed on Dataset-A and Dataset-B.
The performances of our proposed work have been evaluated in terms of accuracy (\%) and some evaluation metrics such as precision (PR), recall (RE) and F-score (harmonic mean of precision and recall) values obtained from the confusion matrix. These values are calculated by the indicators as TP (True Positive), FP (False Positive), FN (False Negative), and TN (True Negative) shown in Eqs. \ref{PR}-\ref{F-score}. \\
\begin{equation}\label{PR}
    PR=\frac{TP}{TP+FP} 
\end{equation}
\begin{equation}\label{RE}
RE=\frac{TP}{TP+FN}
\end{equation}
\begin{equation}\label{F-score}
F-score=\frac{2\times PR \times RE}{PR+RE}
\end{equation}

\hspace{-0.5 cm} In our experiment, hand gesture classification accuracy is measured by the following formula: \\ 

\hspace{-0.5cm} Accuracy = $\frac{\text{Number of correctly classified gesture samples}}{\text{Total number of gesture samples in testing sets}} $ \\

\begin{figure*}
 \centering
\includegraphics[width=12cm ,height=24 cm]{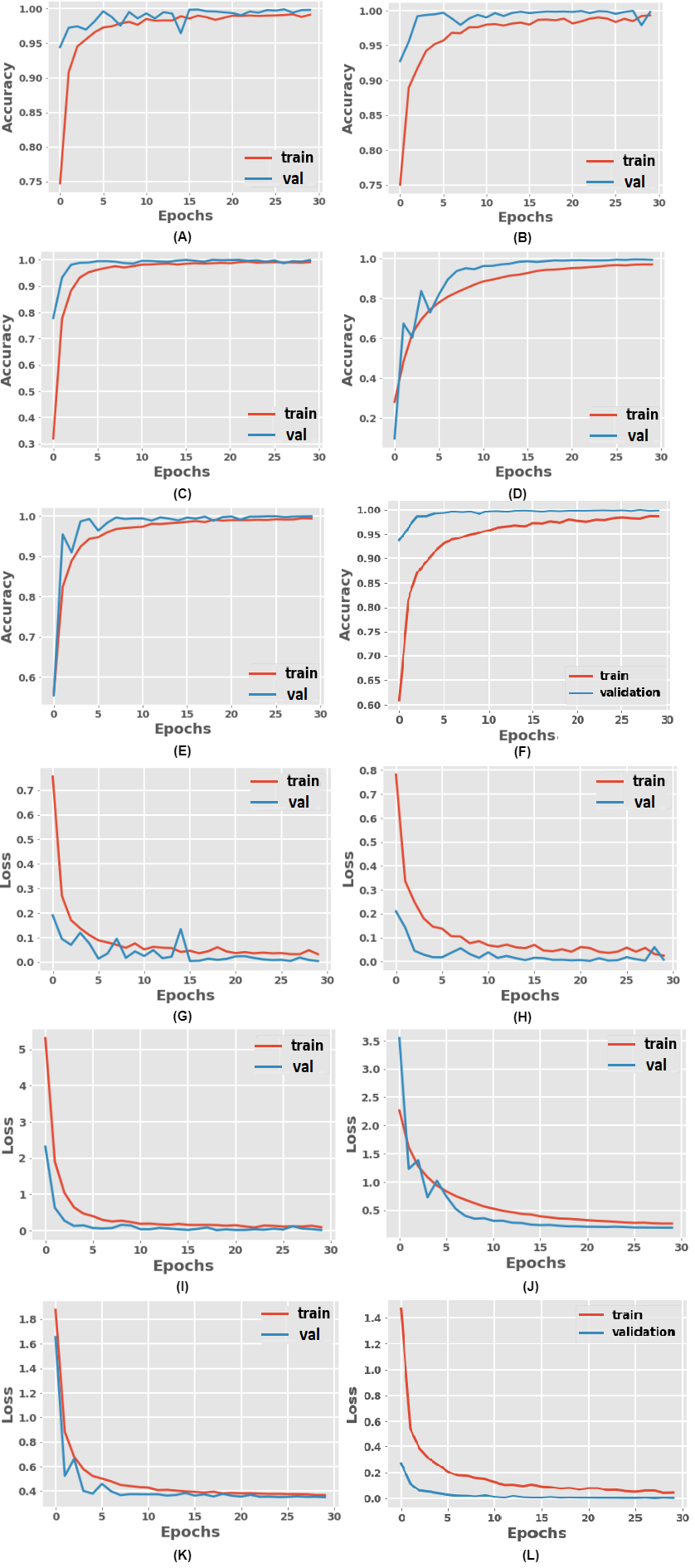}
\caption{\label{dataset-A} Accuracy (training and validation) versus epoch and loss (training and validation) versus epoch plots for each model: Graphs A to F show the accuracy (training and validation) plots of each CNN model and Vision Transformer (VGG16, VGG19, Inception-V1, ResNet50, ResNet101, and ViT respectively). G to L show the loss plot for each model for Dataset-A.}
\end{figure*}

\paragraph{\textbf{Experiment-1:}}\label{experiment1}
This experiment solves ten-gesture class classification tasks based on five pre-trained CNN models over Dataset-A with transfer learning, and vision transformer approach. In Table \ref{CNN1-compare1}, the results of considered five CNN models and ViT over Dataset-A, have been reported. Here, it is observed that the Inception-V1 model has earned maximum classification accuracy compared to other CNN classifiers. In Fig. \ref{dataset-A}, it is seen that the performance of the Inception-V1 model is better than the other models as it has reached minimum loss (training and validation) compared to other models. In Fig. \ref{dataset-A}, it is noticed that there is a lot of oscillation occurring in other models, but the Inception-V1 model
appears to have less fluctuation. It is observed in Table \ref{CNN1-compare1} that the Inception-V1 model has a faster training time (sec)
compared to other considered models (four CNN models and ViT) and provides better accuracy (\%).
The performance comparison of our proposed work (with the best-selected model, i.e., Inception-V1), with other state-of-the-art schemes on Dataset-A has been illustrated in Table \ref{CNN1-compare2} in terms of class-wise accuracy values and the mean value. From comparative analysis (shown in Table \ref{CNN1-compare2}), we can say that our proposed framework (detection + segmentation + morphological transformation + Inception-V1 model) has outperformed the proposed scheme presented by \cite{mantecon2016hand} on Dataset-A in terms of accuracy results (class-wise accuracy). We have also conducted a statistical analysis to estimate the significant difference between the mean of the accuracy values by using one-sample t-test. The entire statistical testing is shown in Appendix Section \ref{appendix-2}.

\begin{table*}[!ht]
\caption{\label{CNN1-compare1}Comparison results among various pre-trained CNN models and ViT on \textbf{Dataset-A} and \textbf{Dataset-B} in terms of training time (sec) and validation accuracy values.}

\begin{center}
\centering
\resizebox*{1\textwidth}{!}{

\begin{tabular}{|ccc|cc|}
\hline
\multicolumn{5}{|c|}{\textbf{Dataset-A}}                                                                                                                                                                                                                                                                                                                                                               \\ \hline

\multicolumn{3}{|c|}{{Input size: $(64 \times 64)$}}                                                                                                                                                                                                                                             & \multicolumn{2}{c|}{{Input size: $(128 \times 128)$}}                 \\ \hline
\multicolumn{1}{|c|}{Model name}                                                                       & \multicolumn{1}{c|}{Training time (sec)} & Validation accuracy (\%)                                                                                                                     & \multicolumn{1}{c|}{Training time (sec)} & Validation accuracy (\%) \\ \hline
\multicolumn{1}{|c|}{VGG16}                                                                            & \multicolumn{1}{c|}{540}                 & 99.67                                                                                                                                        & \multicolumn{1}{c|}{1290}                & 99.60                    \\ \hline
\multicolumn{1}{|c|}{VGG19}                                                                            & \multicolumn{1}{c|}{510}                 & 99.65                                                                                                                                        & \multicolumn{1}{c|}{1500}                & 99.58                    \\ \hline
\multicolumn{1}{|c|}{Inception-V1}                                                                     & \multicolumn{1}{c|}{450}                 & 99.75                                                                                                                                        & \multicolumn{1}{c|}{630}                 & 99.70                    \\ \hline
\multicolumn{1}{|c|}{ResNet50}                                                                         & \multicolumn{1}{c|}{660}                 & 99.60                                                                                                                                        & \multicolumn{1}{c|}{1050}                & 99.58                    \\ \hline
\multicolumn{1}{|c|}{ResNet101}                                                                        & \multicolumn{1}{c|}{870}                 & 99.70                                                                                                                                        & \multicolumn{1}{c|}{1680}                & 99.65                    \\ \hline
\hline
\multicolumn{1}{|c|}{ViT}                                                                        & \multicolumn{1}{c|}{480}                 & 99.70                                                                                                                                       & \multicolumn{1}{c|}{1680}                & 99.65                    \\ \hline
 
\multicolumn{5}{|c|}{\textbf{Dataset-B}}                                                                                                                                                                                                                                                                                                                                                               \\ \hline

\multicolumn{1}{|c|}{VGG16}                                                                            & \multicolumn{1}{c|}{550}                 & 99.25                                                                                                                                        & \multicolumn{1}{c|}{1950}                & 99.20                    \\ \hline
\multicolumn{1}{|c|}{VGG19}                                                                            & \multicolumn{1}{c|}{750}                 & 99.15                                                                                                                                        & \multicolumn{1}{c|}{2400}                & 99.20                    \\ \hline
\multicolumn{1}{|c|}{Inception-V1}                                                                     & \multicolumn{1}{c|}{330}                 & 99.30                                                                                                                                        & \multicolumn{1}{c|}{800}                 & 99.25                    \\ \hline
\multicolumn{1}{|c|}{ResNet50}                                                                         & \multicolumn{1}{c|}{540}                 & 99.05                                                                                                                                        & \multicolumn{1}{c|}{1500}                & 99.00                    \\ \hline
\multicolumn{1}{|c|}{ResNet101}                                                                        & \multicolumn{1}{c|}{900}                 & 98.62                                                                                                                                        & \multicolumn{1}{c|}{2400}                & 98.50                    \\ \hline
\hline
\multicolumn{1}{|c|}{ViT}                                                                        & \multicolumn{1}{c|}{450}                 & 98.60                                                                                                                                        & \multicolumn{1}{c|}{900}                & 98.70                    \\ \hline

\end{tabular}

}
\end{center}

\end{table*}

\begin{figure*}
 \centering
\includegraphics[width=12cm ,height=24 cm]{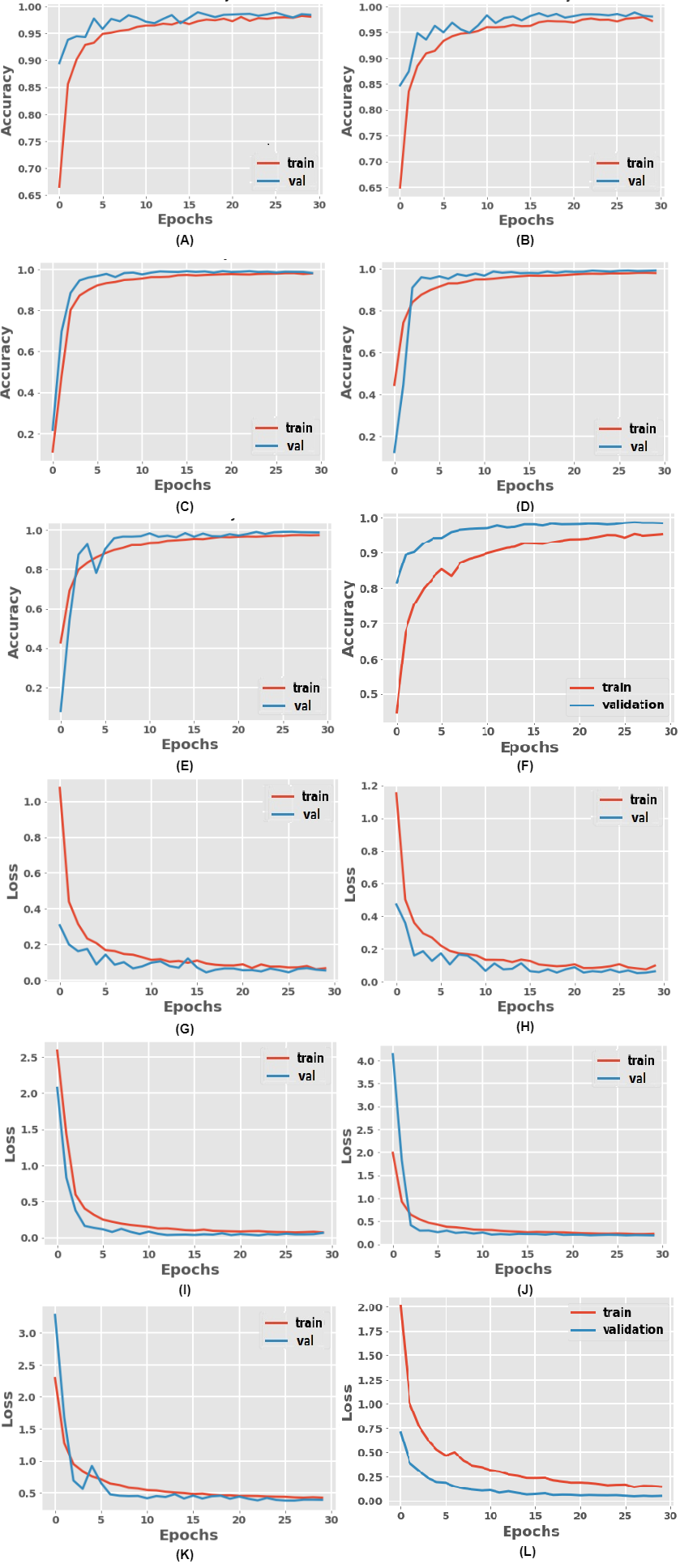}
\caption{\label{dataset-C}  Accuracy (training and validation) versus epoch and loss (training and validation) versus epoch plots for each model: Graphs A to F show the accuracy (training and validation) plots of each CNN model and Vision Transformer (VGG16, VGG19, Inception-V1, ResNet50, ResNet101, and ViT respectively). G to L show the loss plot for each model for Dataset-B.}
\end{figure*}

\paragraph{\textbf{Experiment-2:}}
In this experiment, Dataset-B has been used to solve the 14-gesture class classification task.
Table \ref{CNN1-compare1} reports the validation accuracy results and training time obtained using five pre-trained CNN models and ViT. Table \ref{CNN1-compare1} shows that the Inception-V1 model has achieved the best result compared to other CNN models in terms of training time and validation accuracy. As shown in Fig. \ref{dataset-C}, both plots (accuracy and loss)
C and I show fewer fluctuations, and both the curves (training and validation) have converged. So we can conclude that Inception-V1 has achieved superior results than other pre-trained CNN models and ViT and is further deployed for real-time HCI interface. The performance comparison of our proposed work with the other work proposed presented by \cite{yingxin2016robust} on Dataset-B has been tabulated in Table \ref{CNN3-compare1}. The comparative analysis in Table \ref{CNN3-compare1} exhibits the superiority of our proposed scheme (segmentation, morphological transformation, transfer-learning based CNN models) over other existing scheme \cite{yingxin2016robust} in terms of testing accuracy (\%), precision (\%), recall (\%), and F-score (\%).
\vspace{-0.2cm}

\begin{table}[!ht]
\caption{\label{CNN1-compare2}Comparative analysis of our proposed work (with Inception-V1 model) with existing approach on Dataset-A in terms of accuracy.}

\begin{center}
\centering
\resizebox*{0.50\textwidth}{!}{   
\begin{tabular}{|c|c|c|}
\hline
Gesture class & \begin{tabular}[c]{@{}c@{}}Our work \\ (using Inception-V1)\end{tabular} & \begin{tabular}[c]{@{}c@{}}\cite{mantecon2016hand} Feature descriptor +\\  SVM\end{tabular} \\ \hline
Ok            &   1                                                                         &    0.990                              \\ \hline
Close         &   0.990                                                                          &  1                                 \\ \hline
Palm          &   1                                                                          &     1                             \\ \hline
Thumb         &  1                                                                          &      0.990                            \\ \hline
Index            &   1                                                                          &    1                              \\ \hline
Fist         &     1                                                                        &    0.990                              \\ \hline
Palm move     &   1                                                                          &    1                              \\ \hline
Palm down     &    1                                                                         &    1                              \\ \hline
L             &    1                                                                         &     0.990                             \\ \hline
Fist move         &   0.990                                                                          &    1                              \\ \hline
\textbf{Mean }         &    0.998                                                                         &     0.990                             \\ \hline
\end{tabular}
}
\end{center}
\end{table}

\vspace{-0.2cm}

\begin{table*}[!ht]
\centering

\caption{\label{CNN3-compare1} Performance comparison of our work with other state-of-the-art scheme on Dataset-B.}
\label{CNN3-compare1}
\begin{tabular}{|c|c|C|E|E|E|} 
\hline
Proposed scheme                    & Methods / models                               & Testing Accuracy (\%) & Precision (\%) & Recall (\%) & F-score (\%) \\ 
\hline
Yingxin et al.   \cite{yingxin2016robust}                  & Canny edge detection, CNN                      & 89.17                 & 90.00          & 89.21       & 89.60    \\ 
\hline \hline
\multirow{6}{*}{\textbf{Our Work}} & Gestrure detection, Segmentation, VGG16        & 99.20                 & 99.22          & 99.15       & 99.19    \\ 
\cline{2-6}
                                   & Gestrure detection, Segmentation, VGG19        & 99.00                 & 99.10          & 99.03       & 99.07    \\ 
\cline{2-6}
                                   & Gestrure detection, Segmentation, Inception-V1 & 99.25                 & 99.28          & 99.22       & 99.25    \\ 
\cline{2-6}
                                   & Gestrure detection, Segmentation, ResNet50     & 98.35                 & 98.20          & 98.25       & 98.23    \\ 
\cline{2-6}
                                   & Gestrure detection, Segmentation, ResNet101    & 98.15                 & 98.30          & 98.20       & 98.25    \\ 
\cline{2-6}
                                   & Gesture detection, Segmentation,~ ViT          & 98.50                 & 98.40          & 98.30       & 98.35    \\
\hline
\end{tabular}
\end{table*}

\section{\label{interface} Building of human-machine interface}
Multimedia (for example: audio, and video) plays an essential part in our daily life with the advancement of technology. But physically impaired people find it challenging to interact with the systems. In this work, we have built an interactive interface to control three applications in the real-time scenarios by using corresponding gesture commands after obtaining the best CNN model by using five CNN models and ViT (mentioned in sections \ref{transfer_learning}). The steps followed to build the HCI are illustrated in Algorithm \ref{algo:human-machine}.

\begin{algorithm}
  	\caption{Gesture-controlled human-machine interface}\label{algo:human-machine}
  	\begin{algorithmic}
  		\STATEx \textbf{INPUT}: Give predicted gesture as input commands after training by five pre-trained CNN models and vision transformer.
  		 
  		\STATEx \textbf{OUTPUT}: Development of gesture-controlled HCI interface.
  		
  		\STATE step 1: Choose the best model after training, among the others, in terms of accuracy and training time.
  		
  		\STATE step 2: Get the predicted gesture class label.

  		\FOR{k $\gets$ 1 to C}, where C $\gets$ number of applications.
  		
  		\STATE Control each application according to the customized gesture-class (predicted) label.
  		\ENDFOR
  		
  	\end{algorithmic}
  \end{algorithm}

\subsection{\textbf{VLC-player control:}}
\begin{figure*}[!ht]
\includegraphics[width=\textwidth,height=8 cm]{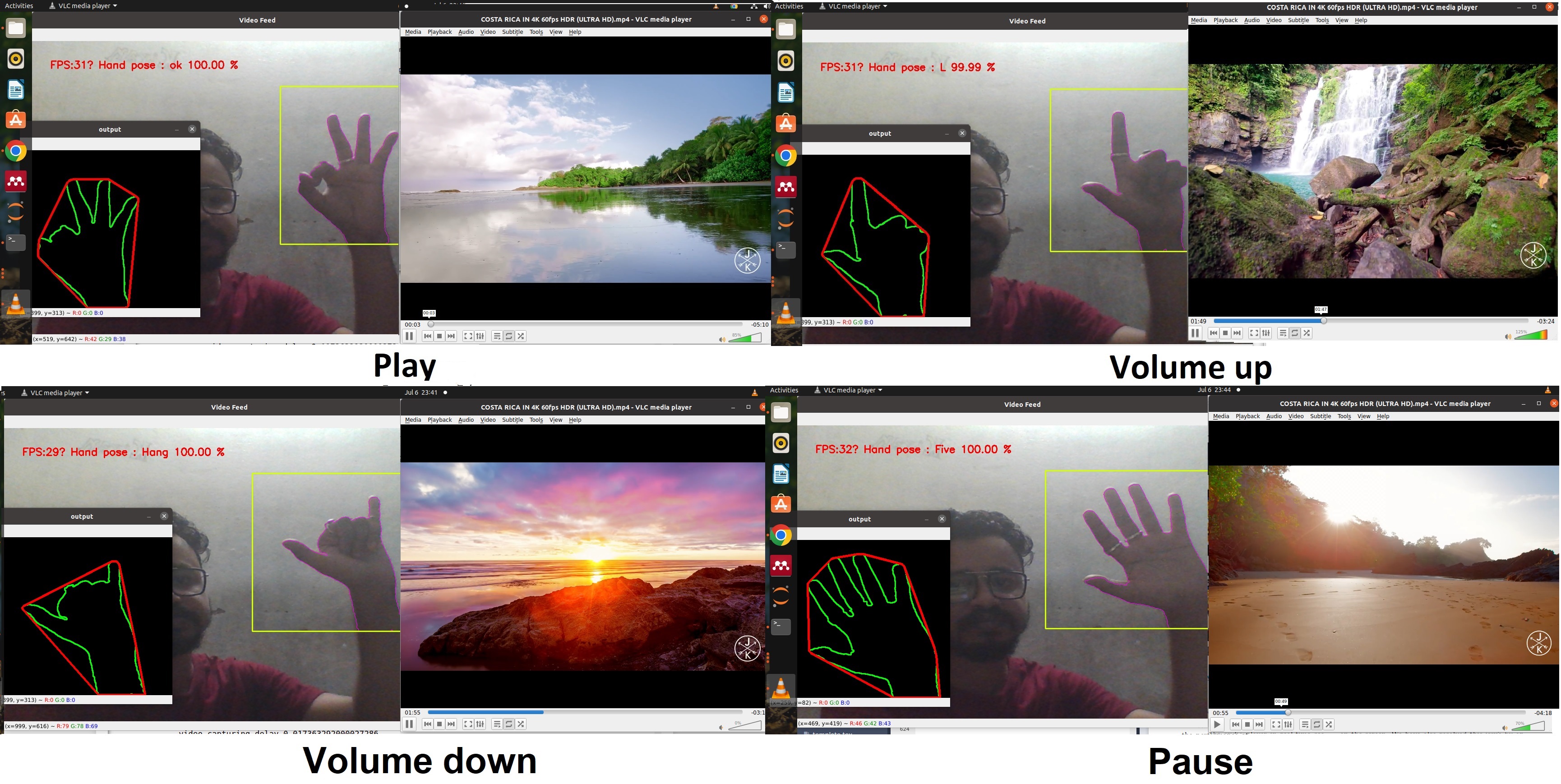}
\caption{\label{vlc_player} VLC player control using gesture commands.}
\end{figure*}

In our work, we have used the `vlc-ctrl' \cite{vlc_ctrl} command-line interface to control various VLC player-related functions, such as play, pause, switch to next video, volume increase/decrease, quit. After obtaining the predicted gesture class label, the keyboard event of the keys is triggered using input gesture commands to control the VLC player. In Table \ref{gesture_controls}, different controls of the VLC player have been tabulated according to various customized gesture labels.
 For example, the gesture `Ok' has been set to start the video. Similarly, gestures `Hang' and `L' for volume up and down, respectively. A demo of this gesture-controlled VLC player has been illustrated in Fig. \ref{vlc_player}. But due to the messy background, this application's performance is sometimes hampered. \\
 The real-time performance analysis of our proposed gesture-command-based VLC-player control has been displayed in Table \ref{vlc_audio_performance}.
 We have also performed every control of the VLC player 10 times with the corresponding gesture-commands to obtain the proper detection rate (\%) along with the response time. In Table \ref{vlc_audio_performance}, we have depicted the number of hits and misses obtained by using our application, followed by calculating the gesture detection rate (\%) and the average response time (in mili second).\\

Gesture detection rate = $\frac{\text{Number of hits}}{\text{Total number of hits and misses}}$ \\

Response time = (time taken to detect gesture $-$ time taken to perform any control of video / audio player with the corresponding gesture commands) 
\\ \\
Average response time= $\frac{\text{Total response time for each function}}{\text{$N$}}$ \\ \\
Here $N$ represents the total number of performance per each control, in our work $N$=10, as we have performed every control of VLC player and audio-player 10 times with the corresponding gesture commands.

\subsection{\textbf{Audio player (Spotify) control:}}
\begin{figure*}[!ht]
\includegraphics[width=\textwidth,height=6.5 cm]{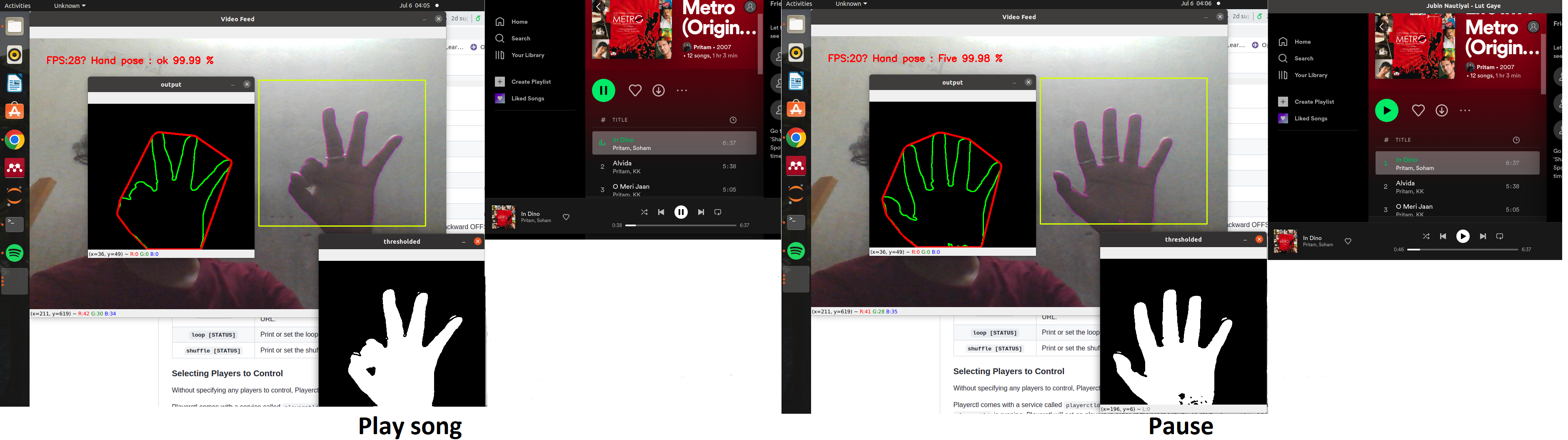}
\caption{\label{audio_player} Audio player (Spotify platform) control using gesture commands.}
\end{figure*}

A summary for controlling different functions of audio-player using various gesture commands has been tabulated in Table \ref{gesture_controls}, whereas Fig. \ref{audio_player} shows the demo of the gesture-controlled audio player. In our work, we have used the `audioplayer' \cite{pypi} python package and `playerctl' command-line library to use different features of the Spotify music player, such as playing a song, pausing, resuming, switching to the next song, and go to the previous music. We have executed every control of the Spotify player (such as: playing, pausing, resuming, etc.) ten times to obtain some valuable results. For instances shown in Table \ref{gesture_controls}, gesture `Ok' has been customized to play the song, similarly, `Five' for pause, `Fist' for resume, `Two' for going to the next song, and `Three' for a switch to the previous music. In Table \ref{vlc_audio_performance}, the performance of the gesture-controlled audio player (Spotify player) in real-time scenario, along with the number of hits, misses, and the average response time (in mili-seconds), has been depicted.

\begin{table}
\centering
\caption{\label{inference_speed}Real-time Performance comparison among all CNN models and Vision Transformer in terms of average inference speed (fps) and time (mili-sec).}
\label{inference_speed}
\begin{tabular}{|c|c|c|} 
\hline
Models       & \begin{tabular}[c]{@{}c@{}}Inference\\~speed (fps)\end{tabular} & \begin{tabular}[c]{@{}c@{}}Inference \\~time (mili-sec)\end{tabular}  \\ 
\hline
VGG16        &      18                                                           & 0.60                                                                 \\ 
\hline
VGG19        &      16                                                           &   0.76                                                               \\ 
\hline
Inception-v1 & 25                                                              &       0.20                                                           \\ 
\hline
ResNet-50    &       18                                                          &   0.50                                                               \\ 
\hline
ResNet-101   &    15                                                             &    0.90                                                             \\ 
\hline \hline
ViT          &   20                                                              & 0.40                                                             \\ 
\hline
\end{tabular}
\end{table}

\begin{table*}[!ht]
\caption{\label{gesture_controls}Desktop applications control by using different gesture commands.}
\begin{center}
    
\begin{tabular}{|cc|cc|cc|}
\hline
\multicolumn{2}{|c|}{\textbf{VLC player}}                       & \multicolumn{2}{c|}{\textbf{Audio player}}           & \multicolumn{2}{c|}{\textbf{2D Super-mario bros}}        \\ \hline
\multicolumn{1}{|c|}{Gesture label} & Action           & \multicolumn{1}{c|}{Gesture label} & Action & \multicolumn{1}{c|}{Gesture label} & Action     \\ \hline
\multicolumn{1}{|c|}{Ok}            & Play             & \multicolumn{1}{c|}{Ok}            & Play   & \multicolumn{1}{c|}{Ok}            & Run        \\ \hline
\multicolumn{1}{|c|}{L}             & Volume Up        & \multicolumn{1}{c|}{Five}          & Pause  & \multicolumn{1}{c|}{Five}          & Jump right \\ \hline
\multicolumn{1}{|c|}{Hang}          & Volume down      & \multicolumn{1}{c|}{Fist}           & Resume & \multicolumn{1}{c|}{Two}           & Hold/Stay  \\ \hline
\multicolumn{1}{|c|}{Close}         & Video quit       & \multicolumn{1}{c|}{Two}             & Go to next song   & \multicolumn{1}{c|}{Four}          & Jump left  \\ \hline
\multicolumn{1}{|c|}{Two}           & Go to next video &

\multicolumn{1}{c|}{Three}              &  Go to previous song & \multicolumn{1}{c|}{Thumb}                &   Jump           \\ \hline
\multicolumn{1}{|c|}{Five}          & Pause            & \multicolumn{1}{c|}{}              &        & \multicolumn{1}{c|}{}             &           \\ \hline
\end{tabular}
\end{center}
\end{table*}

\begin{table*}[!ht]
\caption{\label{vlc_audio_performance}Real-time performance analysis of  gesture-command based VLC and audio player control (10 times).}
\begin{center}
\resizebox*{1\textwidth}{!}{   
\begin{tabular}{|cccccc|}
\hline
\multicolumn{6}{|c|}{Performance analysis of real-time gesture-command based VLC-player application (10 times)}                                                                                                                                  \\ \hline
\multicolumn{1}{|c|}{Action}           & \multicolumn{1}{c|}{Corresponding gestures} & \multicolumn{1}{c|}{Number of hits} & \multicolumn{1}{c|}{Number of  misses} & \multicolumn{1}{c|}{Detection rate (\%)} &  \begin{tabular}[c]{@{}c@{}}Average Response \\ time (ms)\end{tabular} \\ \hline
\multicolumn{1}{|c|}{Play}             & \multicolumn{1}{c|}{Ok}                     & \multicolumn{1}{c|}{10}             & \multicolumn{1}{c|}{0}                 & \multicolumn{1}{c|}{100}                 &      0.35                   \\ \hline
\multicolumn{1}{|c|}{Volume Up}        & \multicolumn{1}{c|}{L}                      & \multicolumn{1}{c|}{8}               & \multicolumn{1}{c|}{2}                  & \multicolumn{1}{c|}{80}                    &    0.83                     \\ \hline
\multicolumn{1}{|c|}{Volume down}      & \multicolumn{1}{c|}{Hang}                   & \multicolumn{1}{c|}{9}               & \multicolumn{1}{c|}{1}                  & \multicolumn{1}{c|}{90}                    &    0.82                  \\ \hline
\multicolumn{1}{|c|}{Quit}             & \multicolumn{1}{c|}{Close}                  & \multicolumn{1}{c|}{10}               & \multicolumn{1}{c|}{0}                  & \multicolumn{1}{c|}{100}                    &       0.38                   \\ \hline
\multicolumn{1}{|c|}{Go to next video} & \multicolumn{1}{c|}{Two}                    & \multicolumn{1}{c|}{7}               & \multicolumn{1}{c|}{3}                  & \multicolumn{1}{c|}{70}                    &      0.38                    \\ \hline
\multicolumn{1}{|c|}{Pause}            & \multicolumn{1}{c|}{Five}                   & \multicolumn{1}{c|}{9}               & \multicolumn{1}{c|}{1}                  & \multicolumn{1}{c|}{90}                    &   0.40                      \\ \hline \hline
\multicolumn{6}{|c|}{Performance analysis of audio-player (10 times) based on gesture command  in real-time scenario}                                                                                                                     \\ \hline
\multicolumn{1}{|c|}{Play}             & \multicolumn{1}{c|}{Ok}                     & \multicolumn{1}{c|}{10}               & \multicolumn{1}{c|}{0}                  & \multicolumn{1}{c|}{100}                    &    0.30                      \\ \hline
\multicolumn{1}{|c|}{Pause}            & \multicolumn{1}{c|}{Five}                   & \multicolumn{1}{c|}{10}               & \multicolumn{1}{c|}{0}                  & \multicolumn{1}{c|}{100}                    &       0.30                   \\ \hline
\multicolumn{1}{|c|}{Resume}           & \multicolumn{1}{c|}{Fist}                    & \multicolumn{1}{c|}{9}               & \multicolumn{1}{c|}{1}                  & \multicolumn{1}{c|}{90}                    &     0.40                     \\ \hline
\multicolumn{1}{|c|}{Go to next song}             & \multicolumn{1}{c|}{Two}                      & \multicolumn{1}{c|}{8}               &     \multicolumn{1}{c|}{2}                  & \multicolumn{1}{c|}{80}                    &                0.35          \\ \hline
\multicolumn{1}{|c|}{Go to previous song}      & \multicolumn{1}{c|}{Three}                  & \multicolumn{1}{c|}{7}               & \multicolumn{1}{c|}{3}                  & \multicolumn{1}{c|}{70}                    &        0.35                  \\ \hline
\end{tabular}
}
\end{center}
\end{table*}

\subsection{\textbf{Super Mario-Bros game control}}

\begin{figure}[!ht]
\includegraphics[width=8 cm,height=7.5 cm]{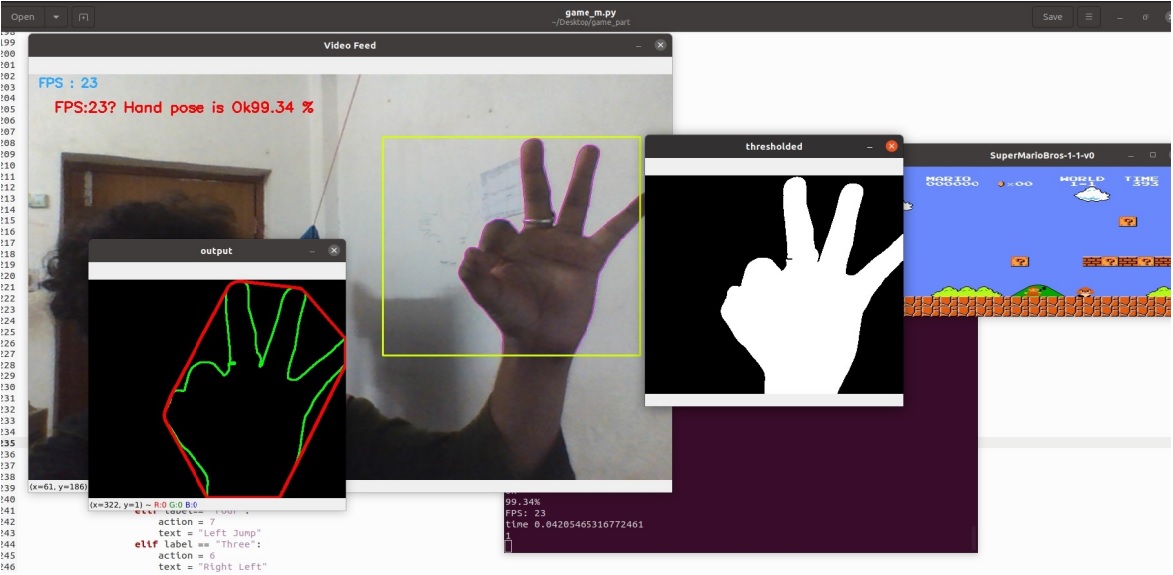}
\caption{\label{mario_bros} 2D Mario-Bros game control using gesture commands.}
\end{figure}

In this section, we have demonstrated the experimental details of the gesture-controlled super-Mario Bros game along with real-time performance analysis. Here we have used the OpenAI Gym environment and imported $`gym - super - mario-bros'$  \cite{gym-super-mario-bros} command-line interface to trigger different controls of this game based on various gesture commands. For example, the gesture `Ok' has been set for the run of Mario character; similarly, gestures `Four' and `Five' have been customized for jumping to the left and right direction, respectively, and gesture `Two' for hold/stay of the Mario. This game's controls are tabulated in Table \ref{gesture_controls}. One Demo of this game is shown in Fig. \ref{mario_bros}.

\begin{figure}[!ht]
\includegraphics[width=7.5 cm,height=7.5 cm]{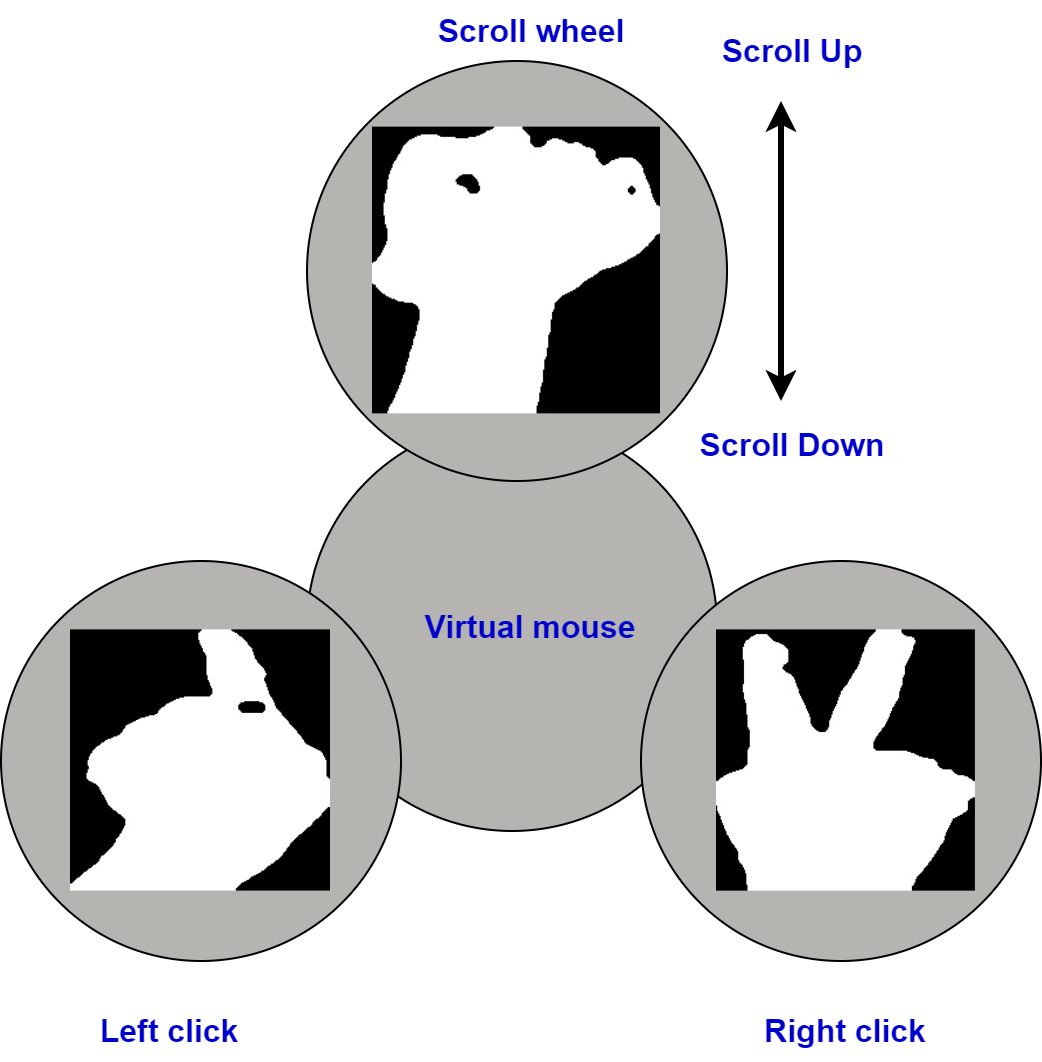}
\caption{\label{virtual_mouse} Schematic diagram of virtual mouse.}
\end{figure}

\begin{figure*}[!ht]
\centering
\includegraphics[width=\textwidth,height=4cm]{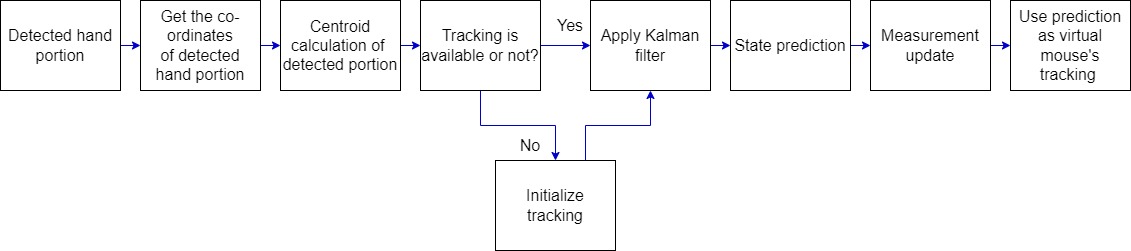}
\caption{\label{kalman_filter} Block diagram for Kalman filter based tracking.}
\end{figure*} 

\begin{figure*}[!ht]
\includegraphics[width=\textwidth,height=7 cm]{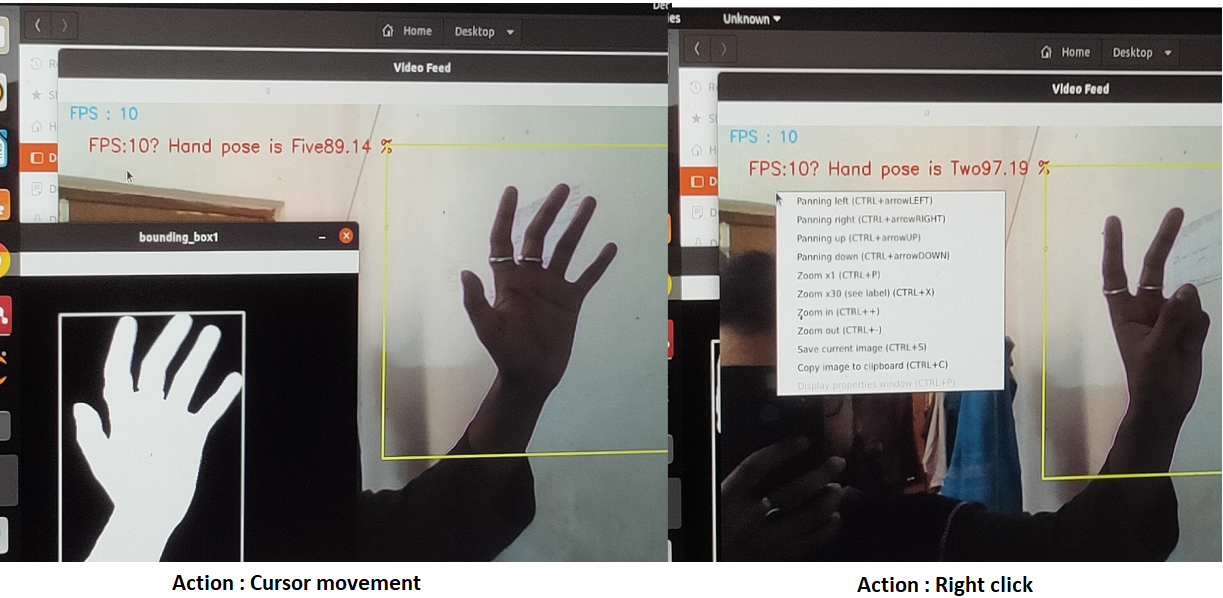}
\caption{\label{virtual_mouse1} Demo of gesture-controlled virtual mouse.}
\end{figure*}

\subsection{Building of virtual mouse}
In order to make the interface more convenient, we have developed a virtual mouse, where main objective is to control various functions of a traditional mouse with the help of predicted gesture commands and a simple web camera instead of a physical mouse device. The main functions of a traditional mouse are click, right-click, double-click, drag, scroll wheel up/down and cursor movement. So in our work, we have mimicked different events of a physical mouse with seven corresponding gesture commands. For instances, the gestures `Two' is set for right click, `Ok' is for double click, `Index' is for left click etc. Fig. \ref{virtual_mouse} shows the schematic diagram of the virtual mouse and one demo of gesture-controlled virtual mouse is depicted in Fig. \ref{virtual_mouse1}. The summary of different controls of virtual mouse according to customized gestures have been shown in Table \ref{mouse_control}. But still there is a problem regarding the smooth movement of the mouse cursor, for example, during slight hand movement, jumping of cursor from one position to another on the screen. We have also resolved this issue by applying the Kalman filtering \cite{asaari2010hand}. In section \ref{mouse-cursor-kalman}, we have shown the performance analysis of improved mouse cursor control using Kalman filter.

\begin{table}[!ht]
\caption{\label{mouse_control}Triggering of mouse events using customized gesture inputs.}
\begin{center}
\begin{tabular}{|c|c|}
\hline
Gesture inputs & Action           \\ \hline
Two            & Right click      \\ \hline
Ok             & Double click     \\ \hline
Index          & Left click       \\ \hline
Five           & Pointer movement \\ \hline
Heavy          & Scroll up        \\ \hline
Hang           & Scroll down      \\ \hline
Palm           & Drag             \\ \hline
\end{tabular}

\end{center}

\end{table}

\subsubsection{\label{mouse-cursor-kalman} Kalman filter based mouse-pointer tracking}

In tracking mouse-cursor movement, first, we need to get the coordinates of the detected hand portion, followed by calculating the centroid position. Centroid calculation is mandatory during hand movements because it is dynamic and changes with respect to time. In Fig. \ref{kalman_filter}, the block diagram of the Kalman filter-based virtual mouse pointer tracking is illustrated. Next, the state space model for the process is described by the movement on both x and y, representing horizontal and vertical coordinates, respectively. We have chosen the pointer movement in a linear way and shown the results according to the movement of the hand from left to right and right to left during the control of the virtual mouse.
We have found that employing the Kalman filter makes the gesture-controlled mouse pointer's movement very smooth, as it prevents the jumping of the cursor through the screen. Fig. \ref{virtual_mouse1}, one demo of Kalman filter-based cursor movement is illustrated. One drawback of exploring the Kalman filter in our work is that it introduces some delay during a real-time scenario, which is acceptable.

\section{\label{discussion}Discussions}
An interactive HCI interface is suggested through a vision-based hand gesture classification system based on deep learning approach. We have performed several experiments on two datasets (details description in section \ref{dataset_desciption}) to evaluate the potency of our proposed work. We have computed the performance of five pre-trained CNN models and ViT approach for input size $(64 \times 64)$ and $(128 \times 128)$. The results have been illustrated in Table \ref{CNN1-compare1}. It is observed that the Inception-V1 model has outperformed the other CNN models and ViT model due to having an Inception module, where $(1 \times 1)$ convolution was used for the dimension reduction purpose. Simultaneously, using this architecture reduces the computation cost, and training time becomes faster. It is also seen in Table \ref{CNN1-compare1} that the time taken to train with the images of dimensions $(64 \times 64)$ pixel is less than the time taken to train the images of size $(128 \times 128)$. However, training with smaller images leads to a huge reduction in computational cost. So, during test-time also, the images are resized into the aspect ratio of $(64 \times 64)$. The performance analysis of our all six used models (five pre-trained CNN models and ViT) in the real-time scenario is tabulated in Table \ref{inference_speed}. It is noticed in Table \ref{inference_speed} that the average inference speed by using the Inception-V1 model and ViT is higher than the other considered models, but Inception-V1 exhibits slightly better performance compared to ViT model, so during real-time inference task, we have considered Inception-V1 model for further deployment. In the last part of this work, we have demonstrated our developed HCI and the gesture-controlled virtual mouse to control different applications (VLC player, Spotify music player, and 2D-Super-Marios game) using custom gesture commands. Table \ref{vlc_audio_performance} depicts the performance analysis for each application in the real-time scenario. The advantages of our proposed system are as follows.

\begin{enumerate}

\item [a)] The average speed of this system is 25 fps (frames per second), which meets the perfect requirements for real-time app control.

\item[b)] This HCI interface is highly beneficial for interacting with desktop applications without touching any mouse or keyboard in a real-time scenario. So this system can be a preferred choice for older or physically disabled people.  
\end{enumerate}
Despite the advantages mentioned above of our proposed system, this system encounters some drawbacks during real-time gesture prediction in the cluttered background because it creates multiple contour regions leading to the segmentation error (failed to segment the gesture portion). Illumination variation and distance issues are also challenges in our proposed system. Another project issue is controlling the higher fps applications, as the average speed of our proposed scheme is 25 fps. Furthermore, using the Kalman filter delays the virtual mouse's performance a bit slow in a real-time scenario. We will explore some effective object detection techniques, such as Faster RCNN \cite{ren2015faster}, EfficientDet \cite{tan2020efficientdet}, etc., to detect the gestures in the messy background and will bring some modifications to the Kalman filtering algorithm to avoid the time delay appears during the real-time performance.

\section{\label {conclusion} Conclusion}
 In this paper, a real-time HCI to control two desktop applications (VLC player and Spotify music player), and one 2D game, along with the development of the virtual mouse using gesture commands, has been depicted. One public and one custom datasets have been considered to validate our proposed approach. We have compared the performance of five pre-trained CNN models and ViT on Dataset-A and Dataset-B. The results show that the Inception-V1 model has achieved better results in terms of training time and validation accuracy (\%). The comparative analysis with other proposed schemes also shows that the Inception-V1 model exhibited superior results and was chosen for the real-time gesture classification task for faster training time (sec) and also better accuracy (\%). We have also discussed the real-time performance analysis of every gesture-controlled application and depicted the average response time (ms) for every control of every application. Moreover, we have applied the Kalman filter to make the motion of the gesture-controlled mouse cursor smoother. In our future work, we will build a robust and efficient HCI interface by connecting other modalities, such as eye-gaze tracking, facial expression, etc., to make the system more convenient and user-friendly.


%
%


\newpage





\bibliographystyle{elsarticle-num}
\bibliography{ref}

\appendix
\section{Appendix}\label{appendix-1}
\subsection{\textbf{VGG16}}
 VGG16 architecture is an one type of variant of VGGNet \cite{simonyan2014very} model, consisting of 13 convolutional layers with kernel sizes of $(3 \times 3)$ and Relu activation function. Each convolution layer is pursued by max-pooling layer with filter size $(2 \times 2)$. Finally the FC layer is added with softmax activation function to produce the final output class label. In this architecture, the depth of network is increased by adding more convolution and max-pooling layers. This network is trained on large-scale ImageNet \cite{deng2009imagenet} dataset, ImageNet dataset  consists of millions of images and having more than 20,000 class labels developed for large scale visual recognition challenge. VGG16 has reported test accuracy of 92.7\% in the ILSVRC-2012 challenge.

\subsection{\textbf{VGG19}}
VGG19 is also a variant of VGGNet network \cite{simonyan2014very}, it comprises 16 convolutional layers and three dense layers with kernel /filter size of $(3 
\times 3)$ and then the max-pooling layer is used with filter size $ (2 \times 2)$ . This architecture is pursued by the final FC layer with softmax function to deliver the predicted class label. This model has achieved second rank in ILSVRC-2014 challenge after being trained on ImageNet \cite{deng2009imagenet} dataset. This model has an input size of $(224 \times 224)$.

\subsection{\textbf{Inception-V1}}
Inception-V1 or GoogleNet \cite{szegedy2015going} is a powerful CNN architecture with having 22 layers built on the inception module. In this module, the architecture is limited by three independent filters such as $(1 \times 1)$, $(3 \times 3)$ and $(5 \times 5)$. Here in this architecture, $(1 \times 1)$ filter is used before 
$(3 \times 3)$, $(5 \times 5)$ convolutional filters for dimension reduction purpose. This module also includes one max-pooling layer with pool size $(3 \times 3)$. In this module the outputs by using convolutional layers such as $(1 \times 1)$, $(3 \times 3)$ and $(5 \times 5)$ are concatenated and form the inputs for next layer. The last part follows the FC layer with a softmax function to produce the final predicted output. This input of this model is $(224 \times 224)$. This architecture is trained with ImageNet \cite{deng2009imagenet} dataset and has reported top-5 error of 6.67\% in ILSVRC-2014 challenge.

\subsection{\textbf{ResNet50:}} 
Residual neural network \cite{he2016deep} was developed by Microsoft research , This model consists of 50 layers, where 50 stands total number of deep layers, containing 48 convolutional layers, one max-pooling. Finally global average pool layer is connected to the top of the final residual block, which is pursued by the dense layer with softmax activation to generate the final output class. This network has input size of $(224 \times 224)$. The backbone of this architecture is based on residual block. In the case of residual block, the output of one layer is added to a deeper layer in the block, which is also called skip connections or shortcuts. This architecture also reduces the vanishing and exploding gradient problems during training. ResNet50 architecture was trained on the ImageNet dataset \cite{deng2009imagenet} and has achieved a good results in ILSVRC-2014 challenge with an error of 3.57\%.

\subsection{\textbf{ResNet101:}}
 
 ResNet101 model consists of 101 deep layers. Like ResNet50, this architecture is also based on the residual building block. In our experiment, we have loaded the pre-trained version of this architecture, trained on ImageNet dataset \cite{deng2009imagenet} that comprises millions of images. This model's default input image size is $(224 \times 224)$.

\subsection{\textbf{Vision Transformer:}}
 
A standard transformer architecture consists of two components (1) a set of encoder and (2) a decoder. But in the case of ViT , it doesn't require the decoder part as it contains only encoder part. In Vision Transformer, firstly image is split into fixed-sized patches, and each patch is implemented for the patch-embedding phase. In the case of patch embedding, each patch is flattened to produce a one-dimensional vector. After the patch-embedding phase, positional embedding is added with the patches to retain the positional information about the image patches in the sequence. Next, they are moved to the transformer encoder. The transformer encoder \cite{bazi2021vision} comprises two components: (1) a Multi-head self-attention block (MHSA) and (2) MLP (multiple-layer perceptron). Hence MHSA block splits the inputs into several number of heads so that each head can learn different levels of self-attention. Then, the outputs of multiple attention heads are concatenated and delivered to the MLP. Next, the classification task is performed by the MLP layer.

\section{Appendix}\label{appendix-2}
\subsection{\label{statistical} Statistical hypothesis testing}
We have also performed a statistical analysis in order to check statistical significance of our model. In Experiment-1 \ref{experiment1}, we have conducted one sample $t$-test with the help of IBM SPSS statistical analysis tool. 
In case of null hypothesis, we have to assume that our model is not statistically significant. \\
To obtain the value of $t$, the following formula is used: 
\begin{center}
t=$\frac{(\overline{X}-\mu)}{\frac{SD}{\sqrt{k}} } $    \end{center}
\begin{table}[!ht]
\caption{\label{one-sample} One sample statistics.}
\centering
\resizebox*{0.5\textwidth}{!}{   
\begin{tabular}{cccc}
\\ \hline
\begin{tabular}[c]{@{}c@{}}Number of \\  samples\end{tabular} & Mean    & Standard deviation & Standard error mean \\ \hline
10                & 99.8300 & 0.2907             & 0.0919  \\ \hline
\end{tabular}
}
\end{table}
\begin{table*}[!ht]
\caption{\label{one-sample-test} One-sample T-test result.}
\centering
\resizebox*{0.75\textwidth}{!}{   
\begin{tabular}{ccccccc}
\hline
\multicolumn{7}{c}{\textbf{One-sample Test}}                    \\\hline
      &    & \multicolumn{2}{c}{Significance}   & Test value= 99  & \multicolumn{2}{c}{\multirow{2}{*}{\begin{tabular}[c]{@{}c@{}}95\% Confidence interval \\ of the difference\end{tabular}}} \\
      &    & \multicolumn{2}{c}{}               &                 & \multicolumn{2}{c}{}                                                                                                       \\ \hline
t     & df & One-Sided p     & Two-Sided p      & Mean Difference & Lower                                                        & Upper                                                       \\
9.026 & 9  & \textless 0.001 & \textless{}0.001 & 0.8300           & 0.6219                                                       & 1.038  \\ \hline                                                    
\end{tabular}
}
\end{table*}
Where $\overline{X}$ is the mean of samples. \\
$\mu$\ the test value. \\
SD sample standard deviation. \\
k size of samples. \\
To get the value of $\overline{X}$, firstly, we have used the ten-fold-cross-validation strategy using Dataset-1, and have calculated the fold-wise accuracy followed by computing the average (considered as sample mean) of these accuracy values with the Inception-V1 (best-selected model for Experiment-1 \ref{experiment1}) model.\\
Table \ref{one-sample} shows that the sample mean ($\overline{X}$), sample size (k), test value ($\mu$), and the standard deviation (SD) are 99.83, 10, 99, and 0.2907, respectively, and the entire statistical analysis of one-sample t-test has been demonstrated in Table \ref{one-sample-test}.\\
The results in Table \ref{one-sample-test} exhibit that $p$-value  \textless 0.001. Here $p$-value is used for hypothesis testing to determine whether there is evidence to reject the null hypothesis. \\
If $p$ \textless $\alpha$ where, $\alpha$ (confidence level) = 0.05, then null hypothesis is rejected. \\ In Table \ref{one-sample-test},
it is observed that $p$-value is very less than $\alpha$, so the null hypothesis is rejected, and
we can say that there is a statistically significant difference in the mean of the accuracy values.
\end{document}